\documentclass[10pt,twocolumn,letterpaper]{article}

\usepackage{cvpr}              
\newcommand{\PAR}[1]{\vskip4pt \noindent{\bf #1~}}
\usepackage[dvipsnames]{xcolor}

\usepackage{multirow}

\newcommand{\0}{\phantom{0}}
\usepackage{textcomp, gensymb}
\newcommand{\dataset}{METU-VisTIR\xspace}
\newcommand\HUGE{\@setfontsize\Huge{50}{60}}

\definecolor{cvprblue}{rgb}{0.21,0.49,0.74}
\usepackage[pagebackref,breaklinks,colorlinks,citecolor=cvprblue]{hyperref}

\newcommand{\method}{XoFTR\xspace}

\def\paperID{12} 
\def\confName{CVPR}
\def\confYear{2024}

\title{XoFTR: Cross-modal Feature Matching Transformer}

\author{Önder Tuzcuoğlu$^{1,3}$ \quad Aybora Köksal$^{1,3}$ \quad Buğra Sofu$^{4}$ \quad Sinan Kalkan$^{2,3}$ \quad A. Aydın Alatan$^{1,3}$\\
$^{1}$ Dept. of Electrical and Electronics Eng.  $^{2}$ Dept. of Computer Eng. \\
$^{3}$ Center for Image Analysis, Middle East Technical University, Ankara, Turkey\\
$^{4}$ ROKETSAN Inc., Ankara, Turkey\\
{\tt\small $^{1,2,3}$\{tuzcuoglu.onder, aybora, skalkan, alatan\}@metu.edu.tr, $^{4}$bugra.sofu@roketsan.com.tr }
}

\begin{document}
\maketitle 
\begin{abstract}
We introduce, \method, a cross-modal cross-view method for local feature matching between thermal infrared (TIR) and visible images. Unlike visible images, TIR images are less susceptible to adverse lighting and weather conditions but present difficulties in matching due to significant texture and intensity differences. Current hand-crafted and learning-based methods for visible-TIR matching fall short in handling viewpoint, scale, and texture diversities. To address this, \method incorporates masked image modeling pre-training and fine-tuning with pseudo-thermal image augmentation to handle the modality differences. Additionally, we introduce a refined matching pipeline that adjusts for scale discrepancies and enhances match reliability through sub-pixel level refinement. To validate our approach, we collect a comprehensive visible-thermal dataset, and show that our method outperforms existing methods on many benchmarks. Code and dataset at \url{https://github.com/OnderT/XoFTR}.

\end{abstract}
\vspace{-.5cm}    
\section{Introduction}

Matching local features across different views of a 3D scene is a fundamental step for e.g. visual camera localization \cite{brahmbhatt2018geometry, sattler2018benchmarking}, homography estimation \cite{dubrofsky2009homography}, and structure from motion (SfM) \cite{schonberger2016structure}. Matching features between visible-thermal images is a special case in image matching. Unlike visible images, thermal infrared (TIR) images are robust against adverse light and weather conditions such as rain, fog, snow, and night \cite{deshpande2021matching, jiang2021review}. However, visible-TIR image matching faces challenges due to differences in texture characteristics and nonlinear intensity differences between the thermal and visible spectra, stemming from distinct radiation mechanisms: Thermal images depict thermal radiation, while visible images capture reflected light \cite{li2013land}. TIR cameras also typically have lower resolution and field of view \cite{jiang2021review, tosi2022rgb}, affecting matching performance.

\begin{figure}
    \centering
    \includegraphics[width=.8\linewidth]{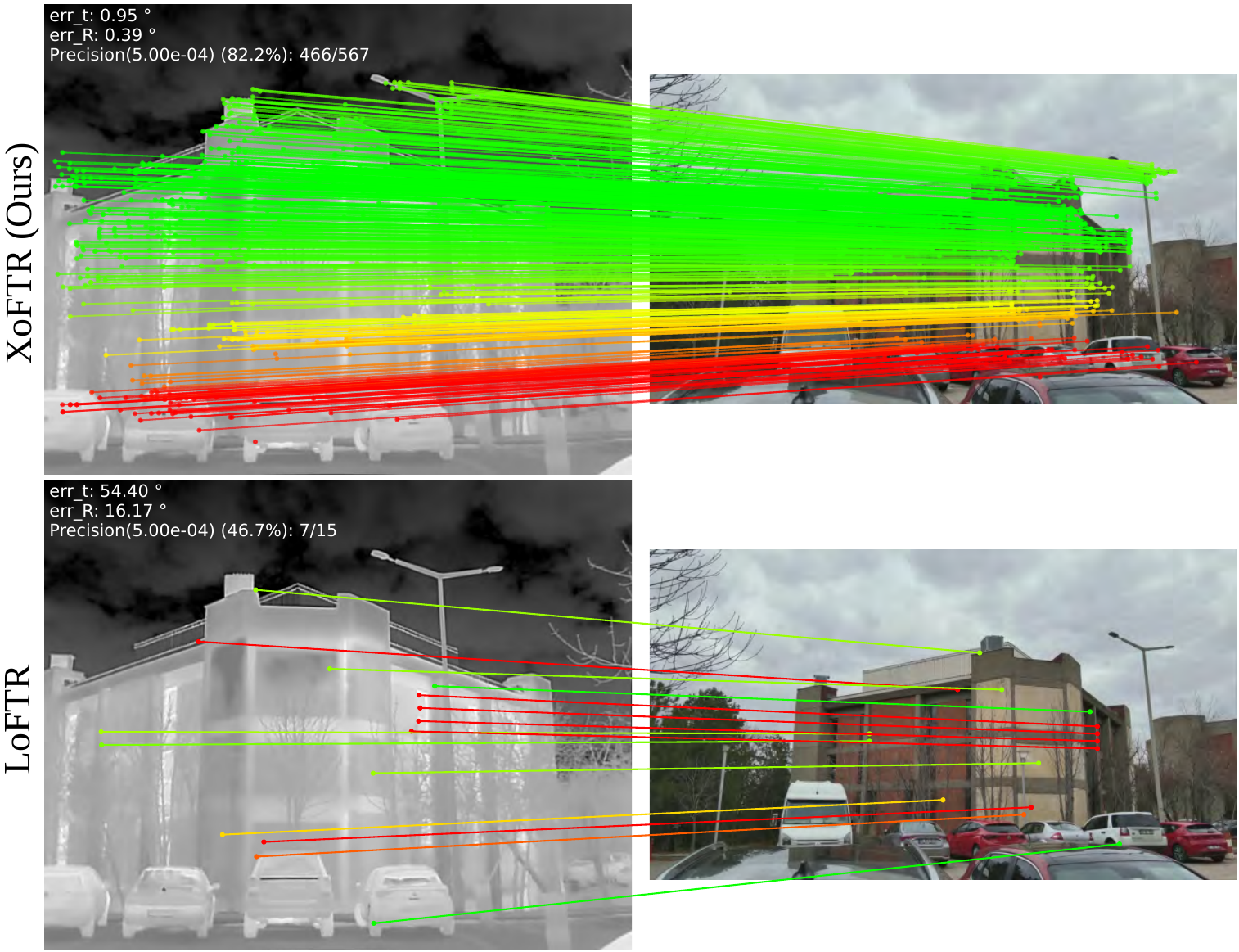}
    \caption{Our \method provides significant improvements over LoFTR \cite{sun2021loftr} on visible and thermal image pairs. Only the inlier matches after RANSAC are shown, and  matches with epipolar error below $5\times 10^{-4}$ are drawn in green.}
    \label{fig:teaser}
    \vspace{-.2cm}
\end{figure}

To match TIR-visible images, many hand-crafted \cite{hu2023multiscale, bilodeau2011visible, han2013visible, li2019rift, hou2024pos} and learning-based  \cite{moreshet2024attention, cui2021cross, chen2023shape, achermann2021multipoint, baruch2021joint} methods have been proposed. Despite the promising results reported, performances across different viewpoints, scales, and poor textures have been sub-optimal. Learning-based methods for visible image matching have addressed many of these challenges but often overlook extreme modality differences in thermal-visible pairs \cite{sun2021loftr, huang2023adaptive, lindenberger2023lightglue, zhu2023pmatch, chen2022aspanformer}.

To address this gap, our study endeavors to extend the methods for visible image matching advancements to the visible-TIR matching problem, choosing the LoFTR network \cite{sun2021loftr} as our baseline model. LoFTR, recognized for its effectiveness in matching through the use of self-attention layers and a correlation-based refinement at the subpixel level, is robust in challenging scenarios, partially also due to the training on the MegaDepth dataset \cite{li2018megadepth}. However, visible-only training sets limit performance in visible-TIR matching. To address this, we propose a two-stage approach adapting (i) masked image modeling (MIM) pre-training and (ii) fine-tuning with augmented images. Inspired by PMatch's MIM \cite{zhu2023pmatch}, our strategy introduces the model to TIR-visible differences, enhancing overall performance. For fine-tuning, we use a robust pseudo-TIR image augmentation method to adapt to modality-induced variations, extending the cosine transform \cite{Fookes2004, Yaman2015}.

As highlighted in prior work \cite{huang2023adaptive}, LoFTR faces challenges with scale differences, often in thermal imaging due to lower resolutions and narrower fields of view. Inspired by AdaMatcher \cite{huang2023adaptive}, we introduce one-to-one and one-to-many matches at 1/8 the original resolution during coarse matching and propose a fine matching pipeline that upscales these matches to 1/2 scale with a customized decoder for both pre-training and fine-tuning, enhancing visible-thermal matching. Re-matching at 1/2 scale filters low-confidence matches, improving reliability of textural structures. Matches are then refined at sub-pixel level using a regression mechanism, preventing a point in one image matching with multiple points in the other.

The absence of a suitable urban visible-thermal benchmark has led us to develop a new dataset for evaluating our method, covering a wide range of viewpoint differences and weather conditions (sunny and cloudy) for comprehensive evaluation. We also evaluate our method's homography estimation performance on publicly available datasets, demonstrating that it surpasses strong baselines, achieving state-of-the-art performance.

\noindent\textbf{Contributions}. Our main contributions are as follows:
\begin{itemize}
  \item We introduce a novel two-stage training approach for visible-thermal image matching, addressing dataset scarcity by leveraging masked image modeling pre-training and fine-tuning with augmented images.
  \item We propose an innovative fine matching pipeline suitable for the pre-training phase of visible-thermal image matching, enabling one-to-many matching and ensuring reliable texture matching at a reduced scale of 1/2.
  \item We curate a novel challenging visible-TIR image matching dataset covering various viewpoint differences and weather conditions. 
  \item Through rigorous experiments, we show that our approach outperforms strong baselines, achieving state-of-the-art results in visible-thermal image matching.
\end{itemize}

\section{Related Work}
\label{sec:related}

\noindent\textbf{Local Feature Matching.}
Detector-based methods for local feature matching can be categorized into handcrafted and learning-based approaches. Following the success of \cite{lowe2004distinctive, bay2006surf, rublee2011orb, calonder2010brief}, handcrafted methods used to be popular before the rise of deep learning based techniques \cite{yi2016lift, tian2017l2, dusmanu2019d2}. 
Deep networks such as Superpoint \cite{detone2018superpoint} and R2D2 \cite{revaud2019r2d2} introduced self-supervised models and joint learning techniques for improved keypoint detection and descriptor discrimination. 
Graph-based methods such as SuperGlue \cite{sarlin2020superglue} and LightGlue \cite{lindenberger2023lightglue} improved efficiency and matching accuracy through graph neural networks and optimized algorithms.

Detector-based approaches often struggle in low-texture areas, a problem which is mitigated in detector-free end-to-end learning-based methods \cite{rocco2018neighbourhood, truong2020glu, truong2020gocor, zhou2021patch2pix, edstedt2023dkm}. The use of Transformer in detector-free matching provided state-of-the-art results \cite{xie2021cotr, sun2021loftr, chen2022aspanformer, huang2023adaptive, zhu2023pmatch}. A prominent example, LoFTR \cite{sun2021loftr}, utilizes a Transformer architecture for local image feature matching, generating matches from coarse to fine, especially in low-texture regions. Other more recent examples include AdaMatcher \cite{huang2023adaptive} and PMatch \cite{zhu2023pmatch}. AdaMatcher tackles large-scale and viewpoint variations with an innovative feature interaction module and adaptive matching for precise patch-level accuracy. PMatch \cite{zhu2023pmatch} redefines dense geometric matching through a novel approach to masked image modeling, a cross-frame transformer, and a unique loss function that improves performance in textureless areas.

While all these methods attain high performances in visible imagery, their application to multimodal visible-thermal pairs is limited. Some studies focus to fill this gap, ranging from handcrafted techniques \cite{hrkac2007infrared, bilodeau2011visible, han2013visible, liu2018novel, liu2018robust} to learning-based solutions \cite{wang2018infrared, baruch2018multimodal, moreshet2024attention, achermann2021multipoint, arar2020unsupervised, cui2021cross}. ReDFeat \cite{deng2022redfeat} re-couples detection and description constraints with a mutual weighting strategy, increasing the training stability and the performance of the features. Shape-Former \cite{chen2023shape} and MIVI \cite{di2023mivi} represent advanced matching methods for multimodal image pairs, emphasizing feature matching and structural consensus.

As summarized in Tab. \ref{table:comparative_summary}, proposed work distinguishes itself from previous approaches by supporting both multimodality and multiview simultaneously. \method is robust across various angles and scales, as well as textures on objects in images of different modalities.

\begin{table}\footnotesize
\centering
\begin{tabular}{c|ccc} \hline
     Method & Year & Multimodal & Multiview \\ \hline
     LoFTR \cite{sun2021loftr} & 2021 & $\times$ & \checkmark \\ 
     DKM \cite{edstedt2023dkm} &  2023 & $\times$ & \checkmark  \\
     Shape-Former \cite{chen2023shape} & 2023 &  \checkmark & $\times$ \\
     MIVI \cite{di2023mivi} & 2023 & \checkmark & $\times$ \\
     AdaMatcher \cite{huang2023adaptive} & 2023 &  $\times$ & \checkmark \\
     PMatch \cite{zhu2023pmatch} & 2023 & $\times$ & \checkmark \\ \hline
     XoFTR  & \textbf{Ours} & \checkmark & \checkmark \\ \hline
\end{tabular}
\caption{A comparative study of our approach with prior work. \label{table:comparative_summary}}
\vspace{-.5cm}
\end{table}

\noindent\textbf{Unsupervised Pre-training in Vision.}
%
%
Following BERT \cite{devlin2018bert} and GPT \cite{radford2018improving} in NLP, unsupervised pre-training (UPT) has become widely used in computer vision, notably with DINO \cite{caron2021emerging}. 
Inspired by context autoencoders \cite{pathak2016context}, denoising autoencoders \cite{Vincent2010}, and masked language modeling as a UPT task in BERT, many studies  \cite{Zhou2022, Bao2022, Zhou2023} has been introduced MIM as a UPT for learning useful representations of images by predicting masked image regions. Approaches using MIM explored different masking and UPT strategies: e.g., the Masked Autoencoder (MAE) \cite{He2022} focuses on partially observed patches, whereas SimMIM \cite{Xie2022} operates by random selection from fully observed patches. 

Despite the unprecedented success of UPT strategies in tasks using RGB imagery, the research on UPT in multi-modal settings has been sparse and only very recent. MultiMAE \cite{Bachmann2023} is one of the few attempts that enhances cross-modal learning by reconstructing masked patches from different modalities, improving task performance without needing specific multi-modal datasets. Complementary Random Masking \cite{Shin2023} targets RGB-Thermal segmentation with unique masking and self-distillation approach, reducing modality dependency. Additionally, a multi-modal transformer \cite{Chen2023} employs masked self-attention for efficient learning with incomplete multi-modal data.

Our presented approach differs from the aforementioned studies on multi-modal UPT by introducing a scheme to adapt the paired pre-training method of PMatch \cite{zhu2023pmatch} to the network structure of LoFTR for Visible to Thermal image applications. This method is adaptable for both pre-training and fine-tuning stages. 

\noindent\textbf{RGB - Thermal Image Conversion.}
%
%
The literature discusses hand-crafted and learning-based methods for RGB to thermal image conversion. One of the simplest yet noteworthy hand-crafted approaches is the cosine transform, introduced by Fookes et al. \cite{Fookes2004} and further explored by Yaman and Kalkan \cite{Yaman2015}. This method is simple and computationally light but produces images lacking true thermal characteristics.

To generate more realistic thermal imagery, several studies have introduced learning-based methods for RGB to TIR, e.g., using Generative Adversarial Network (GAN) based unpaired image to image translation \cite{Kniaz2019, Devaguptapu2019, Zhang2019, Ozkanoglu2022}. The central goal in such approaches is to understand the correlation between RGB an TIR images and simulate thermal image as realistic as possible. 
However, GANs may generate artifacts, if test images differ significantly from training data. To mitigate this, contrastive learning \cite{Park2020} and dual contrastive learning \cite{Han2021} have been proposed. A multi-domain translation network introduced by Lee \cite{Lee2020}, and its edge-preserving modification \cite{Lee2023}, use separate vectors for content and style, aiding in domain translation and preserving edges, respectively, even in the absence of annotated TIR datasets. 

Despite these recent advancements, generative networks for TIR conversion still produce outputs with assumptions and artifacts, leading to pixel inaccuracies \cite{deshpande2021matching}. Therefore, the cosine transform method is preferred for the purposes of this work due to its reliability.

\section{Methods}
\label{sec:methods}
\begin{figure*}[t!]
  \centering
  \includegraphics[width=0.8\linewidth]{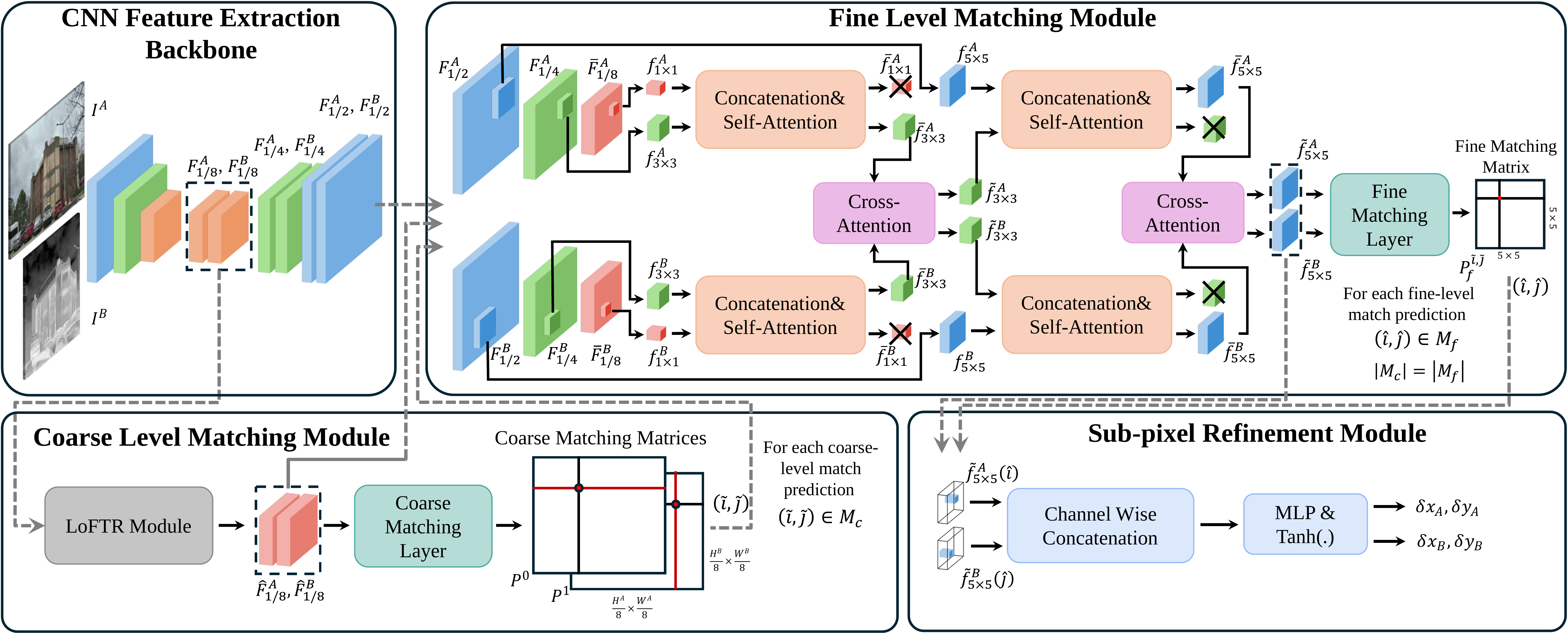}

   \caption{\textbf{Overview of the proposed method.} \method consists of four modules: (1) A CNN backbone which extracts features at scales of $1/8$, $1/4$, and $1/2$. (2) The coarse-level matching module (CLMM), which correlates the features and creates coarse-level match predictions (at $1/8$ scale), allowing one-to-one and one-to-many assignment. (3) The fine-level matching module (FLMM), which re-matches coarse-level match predictions at $1/2$ scale and creates fine-level match predictions, filtering low-confidence matches. (4) The sub-pixel refinement module (SPRM) for refining fine-level match predictions at the sub-pixel level with a regression mechanism, preventing a point in one image from matching with multiple points in the other image.}
   \label{fig:overview}
   \vspace{-.5cm}
\end{figure*}

Our method \method utilizes a ResNet-based \cite{he2016deep} CNN for multi-scale feature extraction from visible and thermal images, integrating three modules: coarse-level and fine-level matching modules, and a sub-pixel refinement module, for precise image match predictions at multiple resolutions. Starting with feature extraction, the approach progresses through coarse and fine-level matching to determine image feature correspondence across scales, and concludes with sub-pixel refinement for accurate match localization. We introduce a new paired masked image modeling (MIM) method of PMatch \cite{zhu2023pmatch} for semi-dense matching, pre-training with real image pairs and fine-tuning with pseudo-thermal images created from visible image datasets through cosine transform augmentation. An overview of the proposed method is presented in Fig. \ref{fig:overview}.

\subsection{Coarse-Level Matching Module}
The coarse-level matching module aims to predict matches at a $1/8$ scale of the original image resolution using coarse-level features ${F_{1/8}^A, F_{1/8}^B}$ derived from the CNN backbone. Unlike the original LoFTR architecture, which employs one-to-one assignment for coarse-level match predictions, we adopt the many-to-one/one-to-many/one-to-one assignment strategy proposed in AdaMatcher \cite{huang2023adaptive}. This approach addresses feature inconsistency caused by large-scale or viewpoint variations common in visible-thermal image matching, eliminating the need for manual methods such as image cropping.

\PAR{LoFTR Module:} We directly adopted the LoFTR module \cite{sun2021loftr}, which consists of linear self- and cross-attention \cite{katharopoulos2020transformers} blocks, to correlate the feature maps $F_{1/8}^A$ and $F_{1/8}^B$, providing refined feature maps denoted as $\hat{F}_{1/8}^A$ and $\hat{F}_{1/8}^B$.

\PAR{Matching Layer:} Given $\hat{F}_{1/8}^A$ and $\hat{F}_{1/8}^B$, firstly, the similarity matrix $\mathcal{S}$ is calculated as:
\begin{equation}\footnotesize
\begin{aligned}
\mathcal{S}(i,j)=\frac{1}{\tau}\cdot\left\langle \operatorname{Linear}(\hat{F}^{A}_{1/8}(i)), \operatorname{Linear}(\hat{F}^{B}_{1/8}(j))\right \rangle,
\end{aligned}
\end{equation}
where $\operatorname{Linear}(\cdot)$ is linear layer, $i$ and $j$ indices of features in feature map $\hat{F}_{1/8}^A$ and $\hat{F}_{1/8}^B$, and $\left \langle \cdot , \cdot \right \rangle$ stands for the inner product. Inspired by \cite{huang2023adaptive}, coarse-level matching probability matrices are obtained by:

\begin{equation}\footnotesize
\begin{aligned}
& \mathcal{P}^{0}(i, j) = \operatorname{Softmax}\left(\mathcal{S}\left(i, \cdot \right)\right)_j , \\ 
& \mathcal{P}^{1}(i, j) = \operatorname{Softmax}\left(\mathcal{S}\left(\cdot, j\right)\right)_i. \\
\end{aligned}
\end{equation}

From the matching probability matrices $\mathcal{P}^{0}$, we select pairs $(i,j)$ as matches when the corresponding confidence values are higher than a threshold value $\theta_c$ and than any other element along its rows. Similarly, for $\mathcal{P}^{1}$, indexes higher than the threshold and other elements in the column were selected as match predictions. We represent coarse-level match predictions as:
\begin{equation}\footnotesize
\begin{aligned}
    \mathcal{M}_{c} & = & \{\left(\tilde{i}, \tilde{j}\right) \mid \mathcal{P}^{0}(\tilde{i}, \tilde{j}) = \max_k \mathcal{P}^{0}(\tilde{i}, k), \mathcal{P}^{0}\left(\tilde{i},\tilde{j}\right) \geq \theta_{c}\} \bigcup \\
    & &  \{\left(\tilde{i}, \tilde{j}\right) \mid \mathcal{P}^{1}(\tilde{i}, \tilde{j}) = \max_k \mathcal{P}^{1}(k, \tilde{j}), \mathcal{P}^{1}\left(\tilde{i},\tilde{j}\right) \geq \theta_{c}\}.
    \label{eq:coarse-matches}
\end{aligned}
\end{equation}

\subsection{Fine-Level Matching Module}
Given the coarse-level match predictions ($\mathcal{M}_{c}$), FLMM is employed to attain matches at the $1/2$ scale of the original image resolution. For this purpose, we designed a customized decoder architecture that permit it to be used both in the pre-training phase and to carry the information processed by the LoFTR module to upper layers, enhancing fine-level visible-thermal matching ability. Furthermore, matches at the 1/2 scale undergo reassessment based on confidence values, enabling the selection of texturally more reliable matches.

In the decoder structure, firstly, we concatenate $\hat{F}_{1/8}^A$ and $F_{1/8}^A$ along the channel dimension. Then, we apply a point-wise convolution, decreasing the channel size to be equal to the channel size of $F_{1/4}^A$, followed by a depth-wise convolution operation with a kernel size of $3 \times 3$. The same procedure is applied to $\hat{F}_{1/8}^B$ and $F_{1/8}^B$. More formally:
\begin{equation}\footnotesize
\begin{aligned}
\bar{F}^*_{1/8} = \operatorname{Conv_{3 \times 3}}\left(\operatorname{Conv_{1 \times 1}}(\hat{F}^*_{1/8} \mathbin\Vert F^*_{1/8})\right),
\label{eq:convs}
\end{aligned}
\end{equation}
where $*$ is either $A$ or $B$.
For each coarse match $(\tilde{i}, \tilde{j})$, we crop pairs of local windows at corresponding locations from $\{\bar{F}^A_{1/8}, \bar{F}^B_{1/8}\}$, $\{F_{1/4}^A, F_{1/4}^B\}$ and $\{F_{1/2}^A, F_{1/2}^B\}$ with sizes $(1 \times 1)$, $(3 \times 3)$ and $(5 \times 5)$ respectively. For one match pair $(\tilde{i}, \tilde{j}) $, these windows are denoted as $\{f^A_{1\times1}, f^B_{1\times1}\}$, $\{f^A_{3\times3}, f^B_{3\times3}\}$ and $\{f^A_{5\times5}, f^B_{5\times5}\}$. If an index $i$ or $j$ is observed more than once in matches, the corresponding window is copied more than once for each pair $(\tilde{i}, \tilde{j})$. Next, to pass information between local window layers, we down-sample the channel dimension of $f^A_{1\times1}$ and concatenate with $f^A_{3\times3}$, and then pass it through a transformer layer with a self-attention. From the output of the transformer, the windows are splinted back and denoted as $\bar{f}^A_{1\times1}$ and $\bar{f}^A_{3\times3}$. These steps can be formulated as:
\begin{equation}\footnotesize
\begin{aligned}
\{\bar{f}^A_{1\times1}, \bar{f}^A_{3\times3}\} = \operatorname{Split}\left(\operatorname{Tr_{self}}\left(\operatorname{Cat}(\operatorname{Down}(f^A_{1\times1}), f^A_{3\times3})\right)\right),
\end{aligned}
\end{equation}
where $\operatorname{Cat}$ and $\operatorname{Split}$ are token-wise concatenation and splitting operation. $\operatorname{Tr_{self}}$ is a transformer layer with self-attention, and $\operatorname{Down}$ denotes downsampling along the channel dimension. The same procedure is applied to $f^B_{1\times1}$ and $f^B_{3\times3}$ as well with outputs $\bar{f}^B_{1\times1}$ and $\bar{f}^B_{3\times3}$.  After this step, to pass information across images, we use another transformer layer with cross-attention between $\bar{f}^A_{3\times3}$ and $\bar{f}^B_{3\times3}$, where the outputs are denoted as $\tilde{f}^A_{3\times3}$ and $\tilde{f}^B_{3\times3}$ as expressed by $\{\tilde{f}^A_{3\times3}, \tilde{f}^B_{3\times3}\} = \operatorname{Tr_{cross}}(\bar{f}^A_{3\times3}, \bar{f}^B_{3\times3})$ where $\operatorname{Tr_{cross}}$ is transformer layer with cross-attention. 

Next, we apply the same steps from the start for window pairs $\{\tilde{f}^A_{3\times3}, f^A_{5\times5}\}$ and $\{\tilde{f}^B_{3\times3}, f^B_{5\times5}\}$, and obtain the outputs $\tilde{f}^A_{5\times5}$ and $\tilde{f}^B_{5\times5}$. For every coarse match prediction $(\tilde{i}, \tilde{j})$, the similarity matrix $\mathcal{S}^{\tilde{i}, \tilde{j}}_f$ between fine-level windows $\tilde{f}^A_{5\times5}$ and $\tilde{f}^B_{5\times5}$ is calculated by $\mathcal{S}^{\tilde{i}, \tilde{j}}_f\left(i, j\right) = \frac{1}{\tau} \cdot \langle\tilde{f}^A_{5\times5}(i), \tilde{f}^B_{5\times5}(j)\rangle$. Then, we employ dual-softmax operation to obtain fine-level matching probability matrix $\mathcal{P}^{\tilde{i}, \tilde{j}}_{f}$: 
\begin{equation}\footnotesize
    \mathcal{P}^{\tilde{i}, \tilde{j}}_{f}(i, j) = \operatorname{Softmax}\left(\mathcal{S}^{\tilde{i}, \tilde{j}}_f\left(i, \cdot \right)\right)_j \cdot \operatorname{Softmax}\left(\mathcal{S}^{\tilde{i}, \tilde{j}}_f\left(\cdot, j\right)\right)_i.
    \label{eq:dual-softmax}
\end{equation}
Finally, for each coarse match prediction $(\tilde{i}, \tilde{j})$, we select the pairs $(\hat{i}, \hat{j})$ for which $\mathcal{P}^{\tilde{i}, \tilde{j}}_{f}(i, j)$ is higher than a threshold of $\theta_f$ and all other elements to obtain fine-level match predictions $\mathcal{M}_f$. As a result for each coarse-level match prediction $(\tilde{i}, \tilde{j})$, we constructed a fine-level match prediction $(\hat{i}, \hat{j})$.

In the employed transformer architectures, we use vanilla attention \cite{vaswani2017attention} and bidirectional attention \cite{bidirectional, lindenberger2023lightglue} for self and cross-attention layers respectively making it more robust to input variations without increasing computational complexity due to small window size and shared query and key projections. Furthermore, we add absolute positional bias to each window feature before sending it to transformer layers to leverage position information effectively. Inspired from \cite{liu2021swinv2, hatamizadeh2023fastervit}, we utilize a 2-layer MLP to embed the absolute 2D token location into the feature dimension.

\subsection{Sub-pixel Refinement Module}
In this module, we convert fine-level match predictions to sub-pixel matches by defining a simple regression mechanism on matches. 
In contrast to \cite{xie2024deepmatcher}, we regress pixel locations for both images. For this purpose, we concatenate the feature vectors of $\tilde{f}^A_{5\times5}$ and $\tilde{f}^B_{5\times5}$ at $(\hat{i}, \hat{j})$ and apply $\operatorname{MLP}$ layer and $\operatorname{Tanh}$ function to jointly regress local sub-pixel coordinates ${\delta_{x_A}, \delta_{y_A}, \delta_{x_B}, \delta_{y_B}}$ as follows: 
\begin{equation}\footnotesize
    \{\delta_{x_A}, \delta_{y_A}, \delta_{x_B}, \delta_{y_B}\} = \operatorname{Tanh}\left(\operatorname{MLP}(\tilde{f}^A_{5\times5}(\hat{i}) \mathbin\Vert \tilde{f}^B_{5\times5}(\hat{j}))\right).
    \label{eq:sub-pixel}
\end{equation}
Then, sub-pixel matches $(\hat{x}_A,\hat{x}_B) \in M_{sub}$ are obtained by summing local sub-pixel coordinates and coordinates of fine-level matches on the images. The sequence of fine-level matching followed by sub-pixel refinement for each coarse match allows us to prevent one point in image $I^A$ from being matched to more than one point in the other image $I^B$ and vice versa. 

\subsection{Masked Image Modeling}
\begin{figure}[t]
  \centering
  \includegraphics[width=0.8\linewidth]{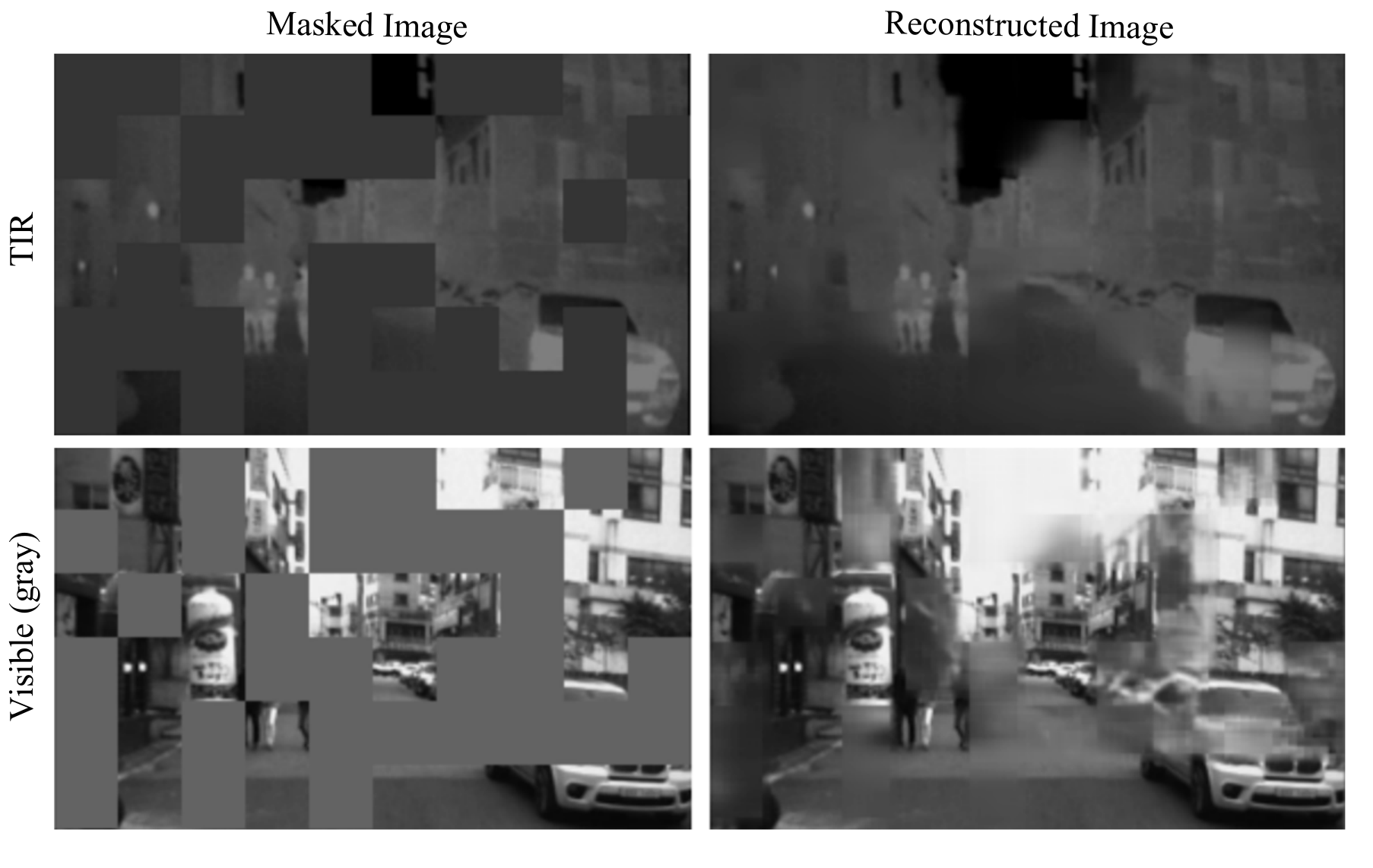}

   \caption{Visualization of reconstructed images using MIM pretext task. Input images are from \cite{hwang2015multispectral}.}
   \label{fig:recons}
   \vspace{-.5cm}
\end{figure}

Before learning to match RGB-IR images, we introduce our model to real multi-modal image pairs with non-linear intensity differences belonging to visible and thermal spectra by utilizing MIM pre-training. Inspired by Pmatch \cite{zhu2023pmatch}, we pre-train our network to reconstruct randomly masked visible-thermal image pairs while conveying pre-trained encoder and decoder layers together to the fine-tuning task. 
\PAR{Masking Strategy:} To use different scale feature maps in the encoder layer used in FLMM, we create the mask in the fine-scale and upscale it up to the original image resolution as in ConvNextv2 \cite{woo2023convnext}. For both images, the masks are generated randomly to cover 50\% of the image with $64\times64$ size patches. Instead of $32\times32$ as in \cite{woo2023convnext, zhu2023pmatch,Xie2022}, we use $64\times64$ patches with a larger input image size of $640\times410$ to enable the network to learn intensity differences in thermal and visible spectra in more detail. We start the masking procedure by applying the binary masks directly on the input images to avoid leakage of masked patches. After passing through the CNN backbone, similar to \cite{Xie2022, zhu2023pmatch, Bao2022}, we use learnable masked token vectors to replace the masked patches on feature maps $F^{A}_{1/8}$, $F^{A}_{1/4}$, $F^{A}_{1/2}$.
\PAR{Decoder:} After the LoFTR module, we directly employ the decoder architecture in FLMM to reconstruct the images. To implement this in practice, for each masked token in coarse scale, we create the local window $\tilde{f}_{5\times5}$ as described in FLMM section and then project it to original image resolution by $\tilde{I}_{10\times10} = \operatorname{Linear(\tilde{f}_{5\times5})}$ to reconstruct the image. In other words, we reconstruct the image using $10\times10$ image windows for each coarse-level masked token. Thanks to the low disparity in the selected dataset \cite{hwang2015multispectral}, the same location of the feature maps for both images   are used with the FLMM layer to benefit from the cross-attention layer. To supervise MIM, we use mean square error (MSE) between the target image and reconstructed image similar to  \cite{woo2023convnext, He2022}. A sample masked sample images and their reconstruction results are shown in Fig. \ref{fig:recons}.

\subsection{Data Augmentation}
\begin{figure}[t]
  \centering
  \includegraphics[width=0.9\linewidth]{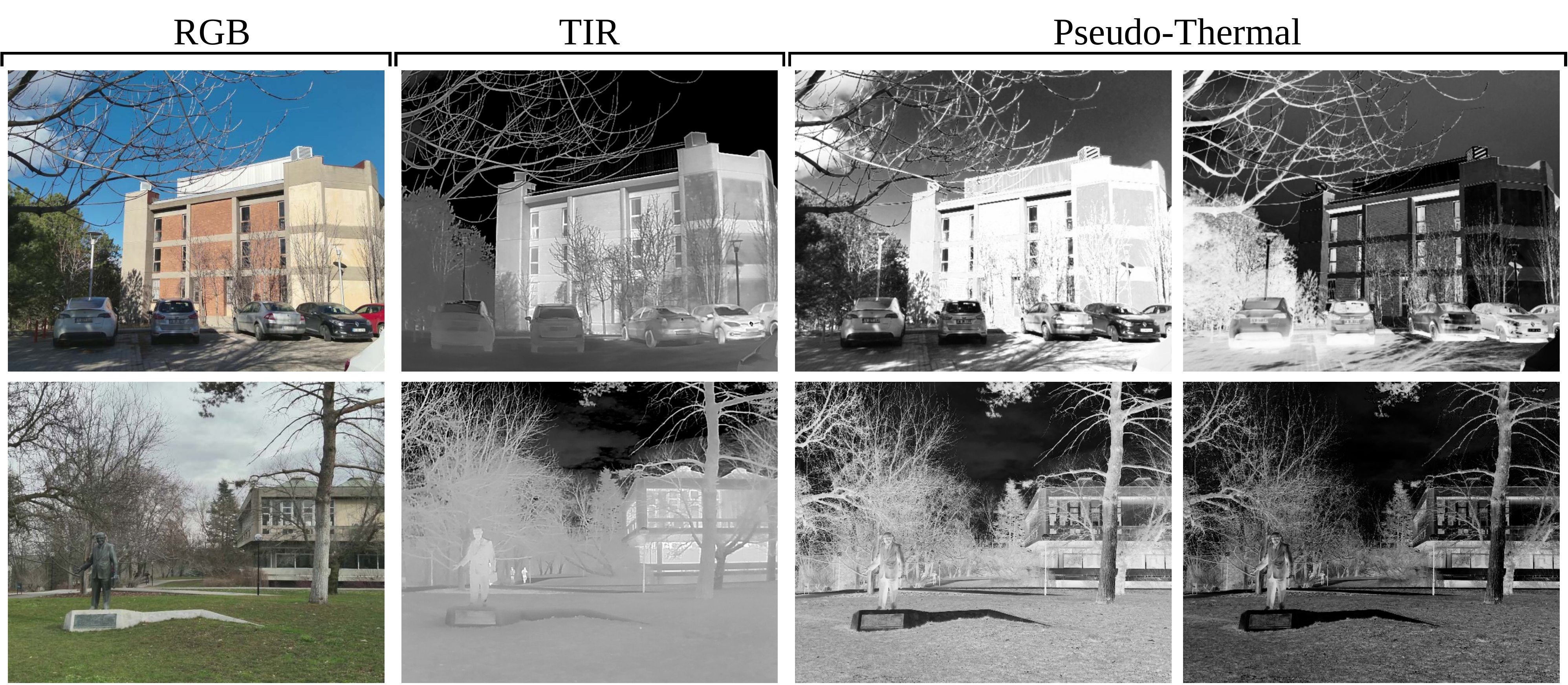}

   \caption{Pseudo-thermal image samples generated with the proposed augmentation method together with real counterparts.}
   \label{fig:pseudo}
   \vspace{-.5cm}
\end{figure}

Due to the lack of urban visible-thermal datasets to be used in image matching, we propose a simple but highly effective image augmentation method to generate pseudo-thermal images from visible images during the fine-tuning stage. To create a pseudo-thermal image, first, we randomly change the hue, saturation, and value intensities of the visible band RGB image. After converting the image to the grayscale, we apply a modified version of the cosine transformation \cite{Yaman2015, Fookes2004} to generate a variety of randomly generated images to represent thermal images with different intensity differences. For the grayscale image $I_{ij} \in [0,1]$ , the randomized cosine transformation is calculated by: 
\begin{equation}\footnotesize
\begin{aligned}
  I_{pseudo} &= \operatorname{Norm}\left(\operatorname{cos}\left(\bar{w}\times\left(I-0.5\right) + \bar{\theta}\right)\right), \text{ for} \\
  \bar{w} &= w_0 + |\alpha_0| \times w_r , \\
  \bar{\theta} &= \pi/2 + \alpha_1 \times \theta_r ,
\end{aligned}
\end{equation}
where $\operatorname{Norm}$ stands for the min-max normalization of the image between 0 and 1.  $\alpha_0$ and $\alpha_1$ are random variables with Normal distribution. $w_0$, $w_r$ and $\theta_r$  are hyper-parameters chosen as $2\pi/3$, $\pi/2$ and $\pi/2$ intuitively. We additionally apply random Gaussian blur operation with kernel size $5\times5$. Some generated pseudo-thermal image samples are shown in Fig. \ref{fig:pseudo} together with real counterparts. By practicing the proposed augmentation method to one of the image pairs during fine-tuning, our network gains endurance against nonlinear intensity variations which is the crucial part for visible-thermal image matching. 
\subsection{Supervision}
Our loss function consists of three components which are coarse-level matching loss, fine-level matching loss and sub-pixel refinement loss.
\PAR{Coarse-Level Matching Loss:} To supervise the matching probability matrices $P^0$ and $P^1$, we apply the Focal Loss ($\operatorname{FL}$) \cite{lin2017focal} following LoFTR \cite{sun2021loftr} and AdaMatcher \cite{huang2023adaptive}: 
\begin{equation}\footnotesize
  \mathcal{L}_c = \left(FL(P^0, \hat{P}) + FL(P^1, \hat{P})\right),
\end{equation}
where $\hat{P}$ is coarse-level ground-truth matching matrix. We obtain ground-truth coarse matches similar to LoFTR \cite{sun2021loftr} but without a mutual nearest neighbor constraint. For this purpose, we create 2D position grids for each image at the $1/8$ scale, and we project these grids to each other using depth maps and camera poses. Then, we assign projected grid points as positive matches using re-projection distances allowing one-to-many and many-to-one assignments.
\PAR{Fine-Level Matching Loss:} Although we select only one point as a match from fine-level windows $\tilde{f}^A_{5\times5}$ and $\tilde{f}^B_{5\times5}$, we supervise all fine-level features correspondences in $\mathcal{P}^{\tilde{i}, \tilde{j}}_{f}$. We define fine-level matching loss as follows: 
\begin{equation}\footnotesize
\begin{aligned}
\mathcal{L}_f = \frac{1}{\left | \mathcal{M}_c \right |}\sum_{(\tilde{i},\tilde{j})\in \mathcal{M}_c} FL\left (\mathcal{P}^{\tilde{i}, \tilde{j}}_{f}, \mathcal{\hat{P}}^{\tilde{i}, \tilde{j}}_{f}\right ),
\end{aligned}
\end{equation}
where $\mathcal{\hat{P}}^{\tilde{i}, \tilde{j}}$ is fine-level ground-truth matching matrix for a coarse-level match $(\tilde{i}, \tilde{j})$. $\mathcal{\hat{P}}^{\tilde{i}, \tilde{j}}$ is calculated at $1/2$ scale similar to coarse-level ground-truth matching matrix with an addition of mutual nearest neighbor constraint allowing only one-to-one matches. 
\PAR{Sub-pixel Refinement Loss:} Inspired by TopicFM+ \cite{giang2023topicfm}, we implemented the symmetric epipolar distance function \cite{hartley2003multiple} to calculate the sub-pixel refinement loss. This approach eliminates the need for explicit ground-truth matching pairs and enables us to supervise both matching coordinates jointly. Given an estimated matching coordinate pair $(\hat{x}_A, \hat{x}_B)$ in normalized image coordinates (in homogeneous form), the sub-pixel refinement loss is defined as:
\begin{equation}\footnotesize
\label{eq:sym_epipolar_loss}
    \mathcal{L}_{sub} = \frac{1}{\left | \mathcal{M}_c \right |}\sum_{(\hat{x}_A, \hat{x}_B)} \| \hat{x}_A^T E \hat{x}_B \|^2 \left( \frac{1}{\|E^T \hat{x}_A\|_{0:2}^2} + \frac{1}{\|E \hat{x}_B\|_{0:2}^2}\right),
\end{equation}
where $E$ is the ground-truth essential matrix obtained using camera poses. 

\noindent\textbf{Overall Loss:} Our total loss is calculated by:
\begin{equation}
 \mathcal{L}_{total} = \lambda_c \mathcal{L}_c  + \lambda_f\mathcal{L}_f + \lambda_{sub}\mathcal{L}_{sub},
\end{equation} where $\lambda_c$, $\lambda_f$ and $\lambda_{sub}$ are hyperparameters chosen as $0.5$, $0.3$ and $10^4$ respectively.

\section{Proposed Dataset}
\label{sec:dataset}
To showcase the effectiveness of our method \method, we introduce \dataset, a novel dataset featuring thermal and visible images captured across six diverse scenes with ground-truth camera poses. Four of the scenes encompass images captured under both cloudy and sunny conditions, while the remaining two scenes exclusively feature cloudy conditions. This diverse dataset facilitates the evaluation of matching algorithms across various challenges, including extreme viewpoint variations and weather-induced changes in lighting and temperature.

\begin{figure}[b]
 \vspace{-.5cm}
  \centering
  \includegraphics[width=0.9\linewidth]{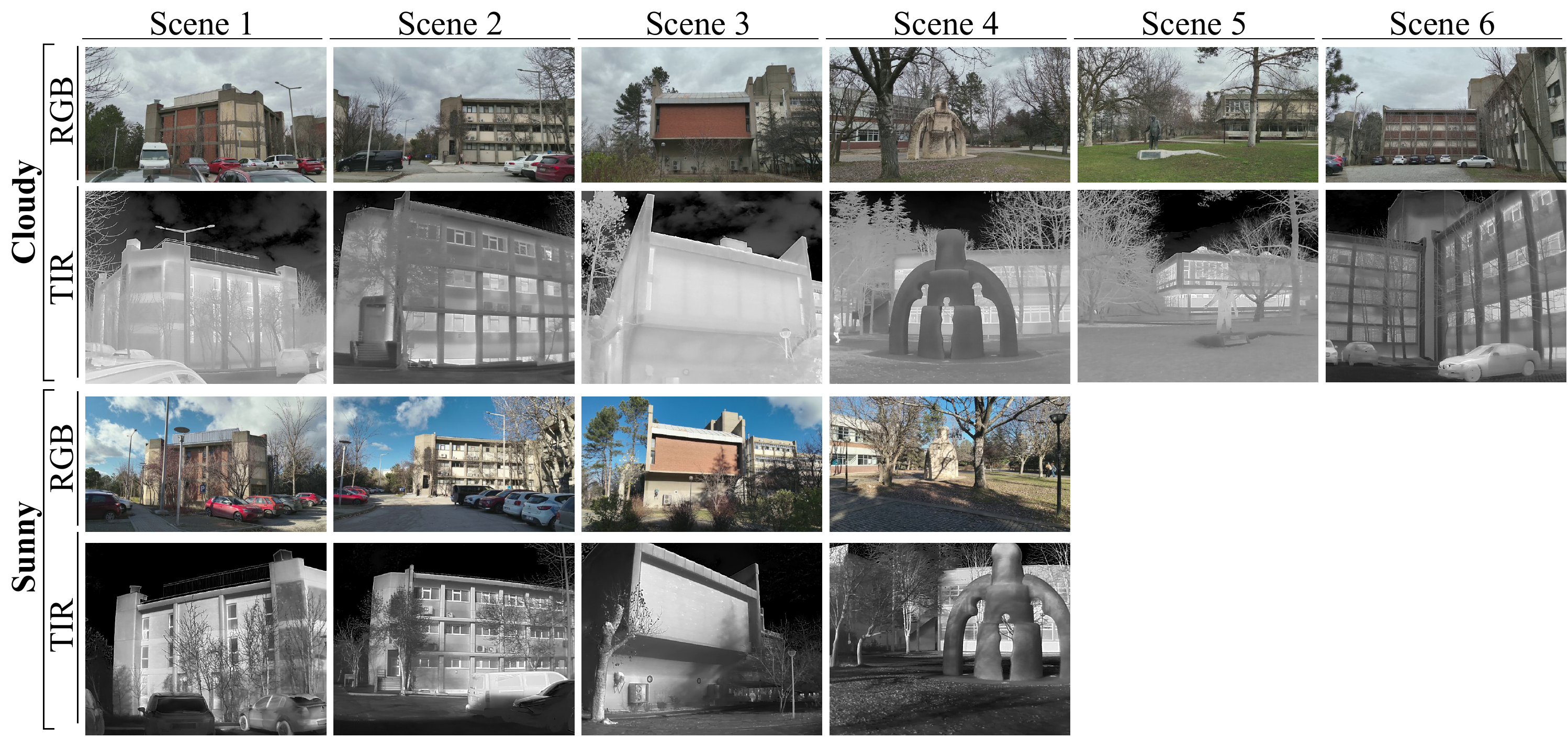}
   \caption{Visualization of some images from our dataset.}
   \label{fig:dataset}
\end{figure}

We captured sequential images of the scenes using cameras of DJI Mavic 3 Thermal drone whose thermal and visible band cameras are positioned closely. The thermal camera boasts a resolution of $640\times512$ pixels, a FOV of $61\degree$, and operates within the wavelength range of 8-14 $\mu$m. Meanwhile, the visible band RGB camera features a resolution of $3840\times2160$ pixels and a FOV of $84\degree$. Ground-truth poses were recovered using offline systems such as COLMAP \cite{schoenberger2016sfm, schoenberger2016mvs} and HLOC \cite{sarlin2019coarse} methods with RGB images. Since the cameras are auto-focus, we obtained GT camera parameters for both cameras using COLMAP in the same modality, supplemented by distortion parameter estimation using a calibration pattern. Although the auto-focus nature of the cameras may result in slight imperfections in the ground truth camera parameters, they are adequate for the purpose of method evaluation. Some images from our dataset are shown in Fig. \ref{fig:dataset}.

\begin{figure*}[htbp]
\centering

\resizebox{.9\textwidth}{!}{
    \begin{tabular}{ccc}

        \Huge LoFTR & \Huge DKM & \Huge XoFTR (\textbf{Ours}) \\
    
        \includegraphics[]{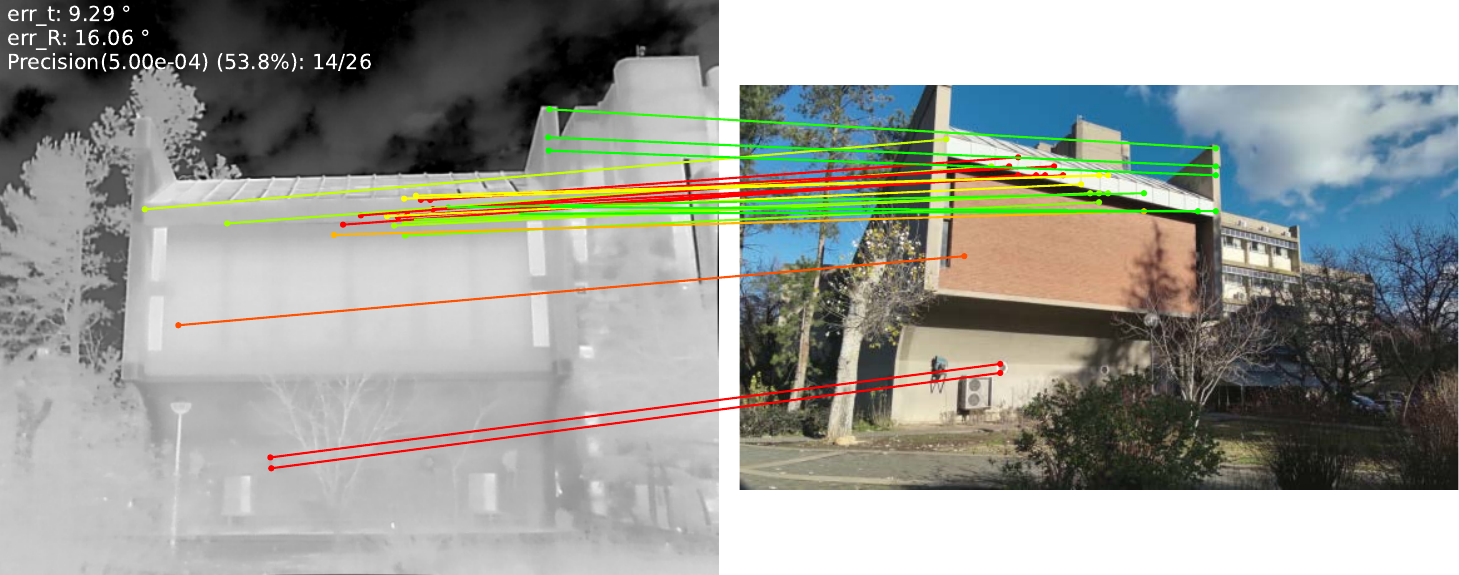} & 
        \includegraphics[]{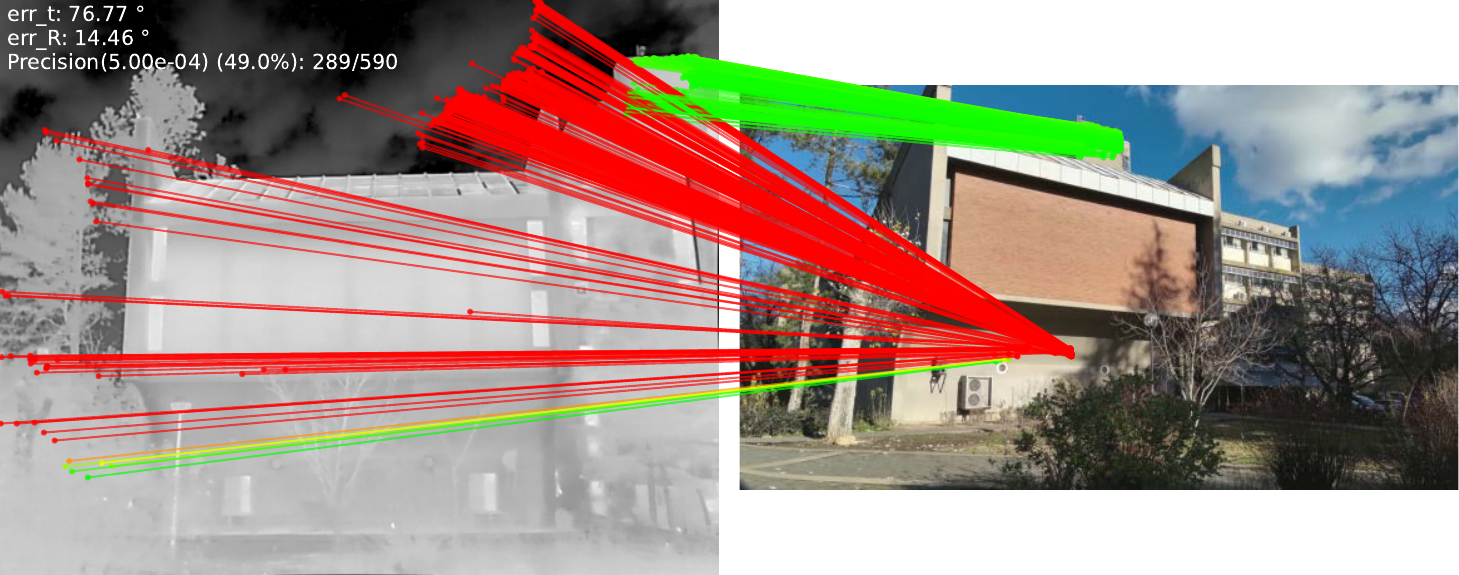} & 
        \includegraphics[]{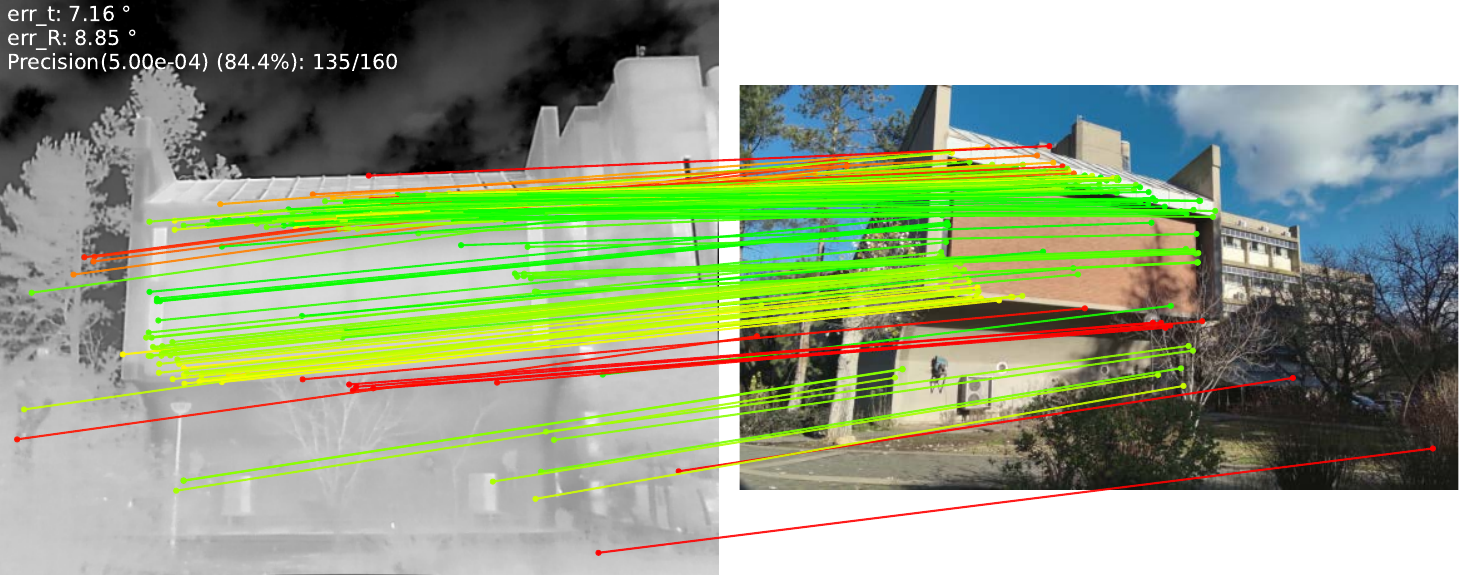} \\
        
        \includegraphics[]{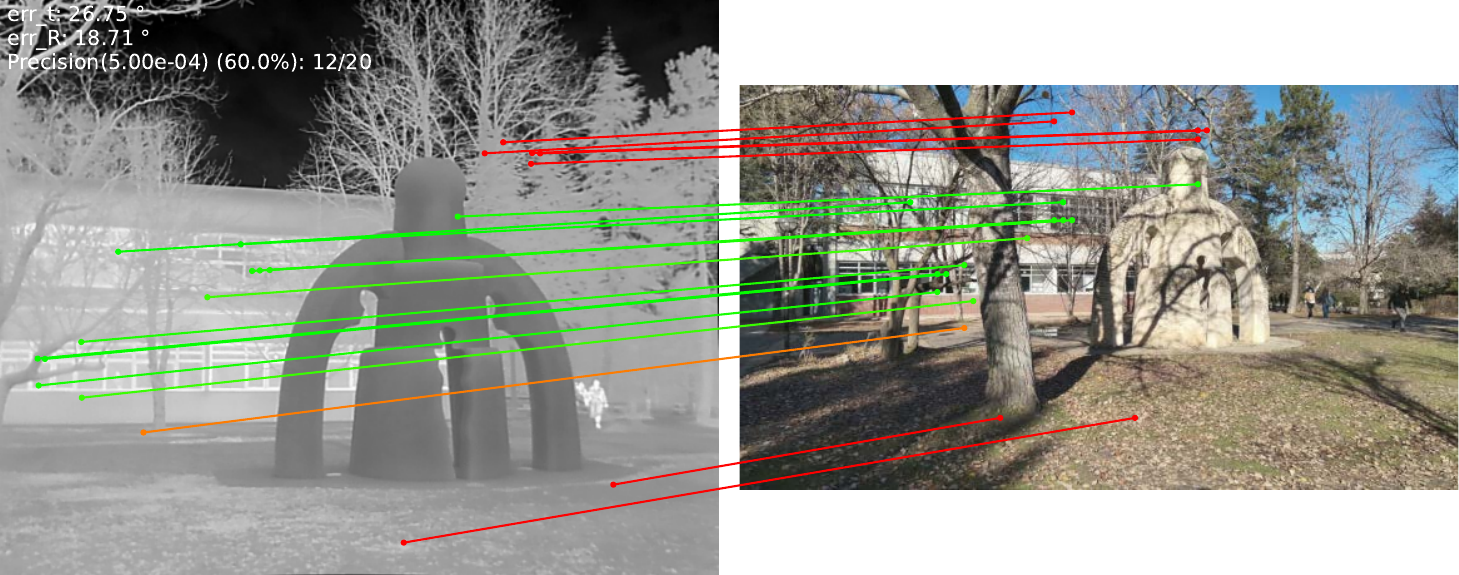} & 
        \includegraphics[]{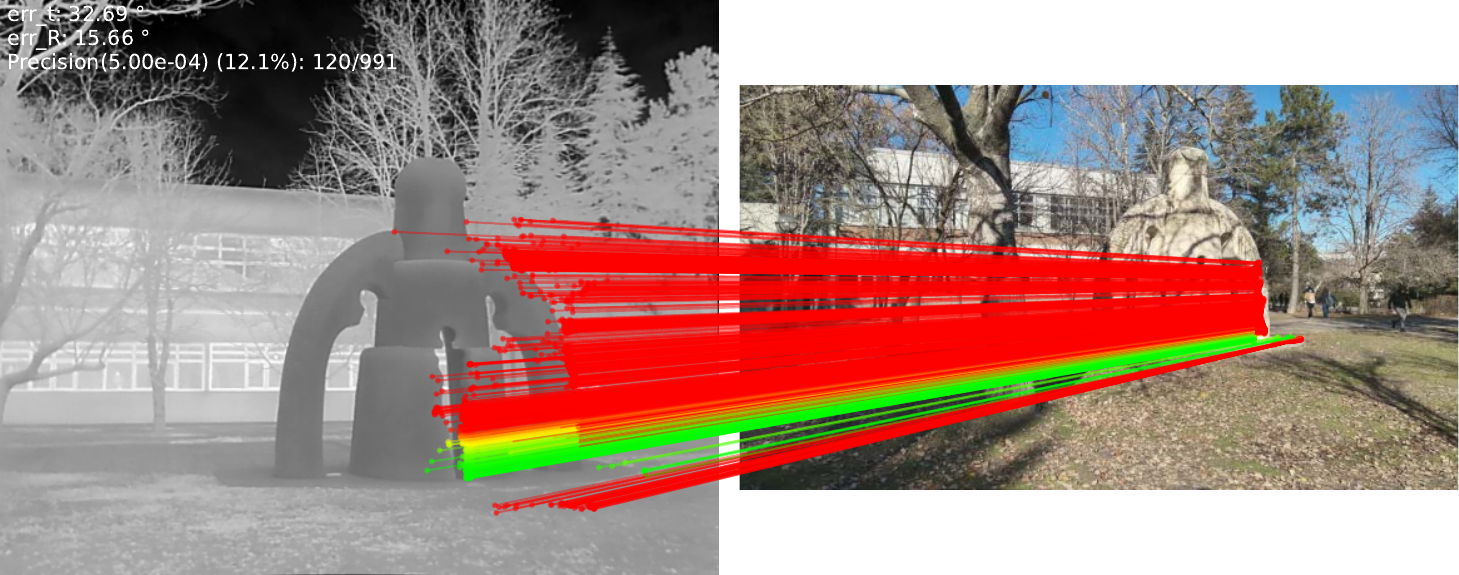} & 
        \includegraphics[]{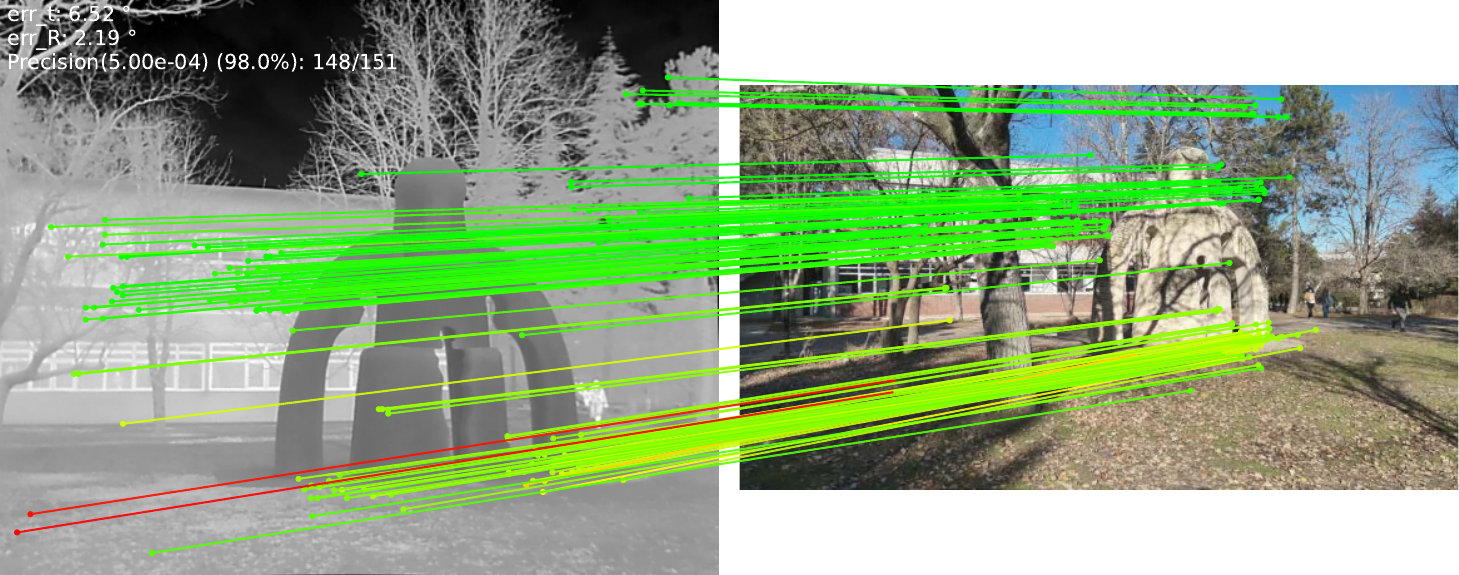} \\

        \includegraphics[]{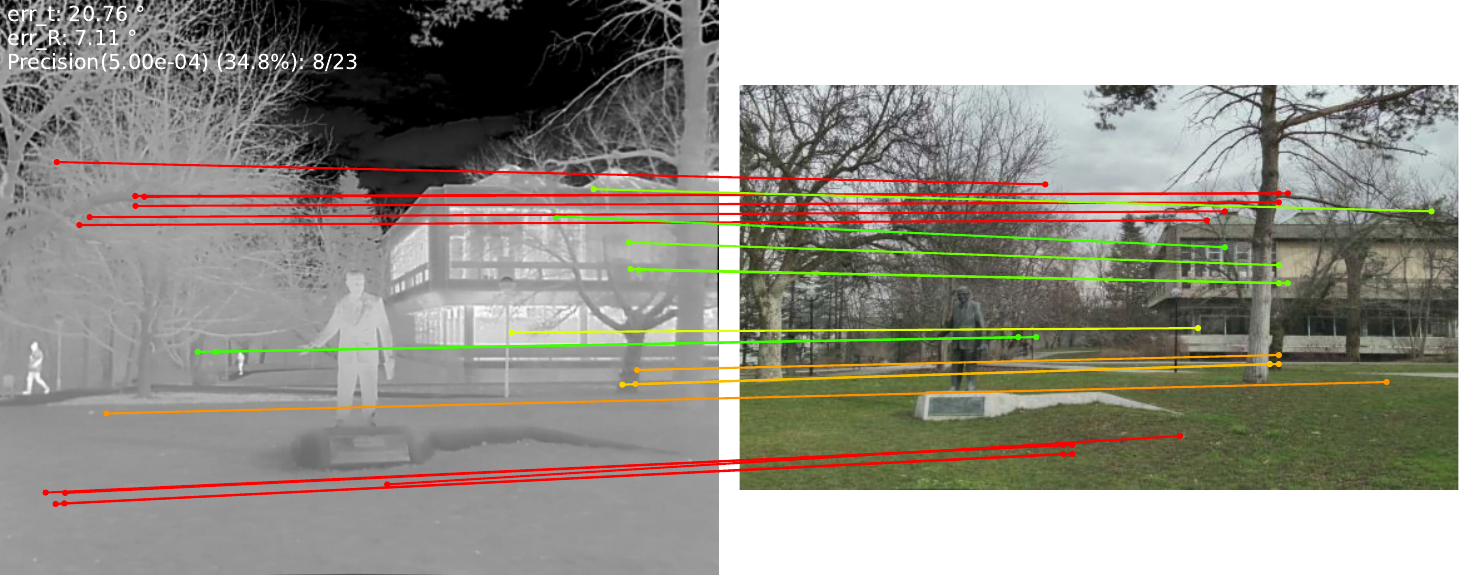} & 
        \includegraphics[]{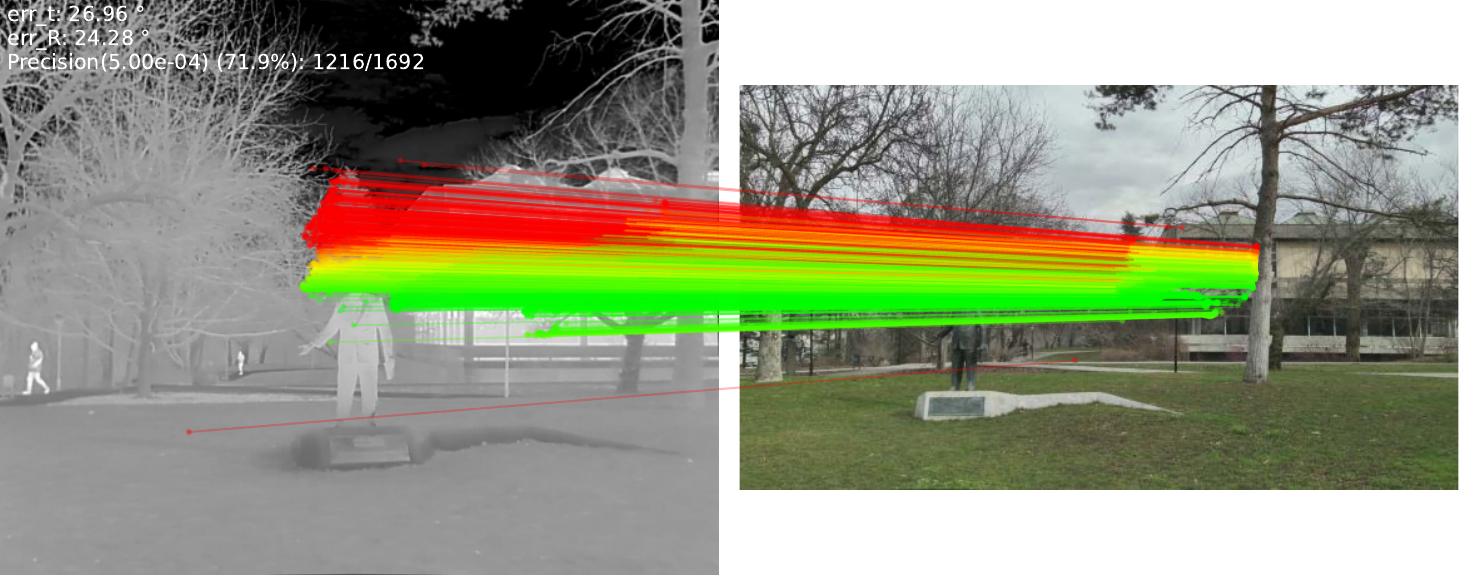} & 
        \includegraphics[]{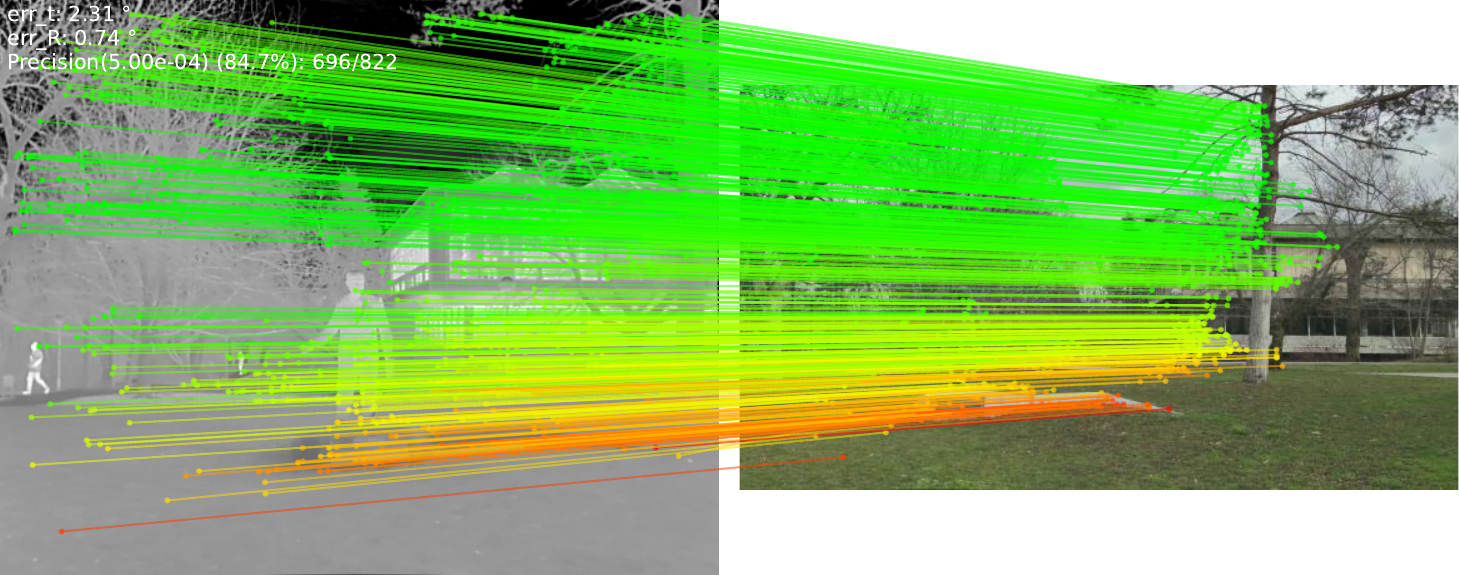} \\
    \end{tabular}
}
\caption{\textbf{Qualitative results for pose estimation.} XoFTR (right column) is compared to DKM and LoFTR in \dataset dataset. Only the inlier matches after RANSAC are shown, and  matches with epipolar error below $5\times 10^{-4}$ are shown in green lines. }
\label{fig:qulitative}
\vspace{-.5cm}
\end{figure*}

We created two benchmark sets from the captured images, totaling 1382 and 1208 image pairs, labeled as cloudy-cloudy and cloudy-sunny. The cloudy-cloudy set consists of thermal and RGB image pairs with corresponding GT camera poses, all captured under cloudy conditions. Conversely, the cloudy-sunny set contains thermal and RGB image pairs with GT poses, capturing one image in sunny and the other in cloudy conditions.

\section{Experiments}
\label{sec:experiment}

\begin{table}[tb]
    \centering
    \scriptsize{
        \setlength\tabcolsep{4.0pt}
\begin{tabular}{clcccc}
    \toprule
    \multirow{2}{*}[-.4em]{Category} & \multicolumn{1}{c}{\multirow{2}{*}[-.4em]{Method}} & \multicolumn{3}{c}{Pose estimation AUC}  \\
    \cmidrule(lr){3-5}
                                               &                                    & @5\degree & @10\degree & @20\degree           \\
    \midrule
    \multirow{3}{1.5cm}[-.0em]{Detector-based} &
     D2-Net~\cite{dusmanu2019d2}+NN             & \02.16    & \06.01      & 12.80     \\
                                               & SP~\cite{detone2018superpoint}+SuperGlue~\cite{sarlin2020superglue}               & \03.90     & \08.75      & 16.35     \\
                                               & SP~\cite{detone2018superpoint}+LightGlue~\cite{lindenberger2023lightglue} &  \01.17     & \03.97      & \09.60 \\
                                               & ReDFeat~\cite{deng2022redfeat} &  \02.36     & \05.45      & 11.26 \\
                                               
    \midrule
    \multirow{5}{1.5cm}{Detector-free}
                                               & \multicolumn{1}{l}{LoFTR~\cite{sun2021loftr}} & \02.63     & \06.55      & 14.11    \\
                                               & \multicolumn{1}{l}{LoFTR-MTV~\cite{liu2022multi}} & \01.54     & \03.89      & \08.80     \\
                                               & \multicolumn{1}{l}{ASpanFormer~\cite{chen2022aspanformer}} & \01.82 & \04.73  & 10.60 \\
                                               & \multicolumn{1}{l}{DKM~\cite{edstedt2023dkm}} & \0\underline{5.79} & \underline{11.47}  & \underline{19.17} \\
                                               & \multicolumn{1}{l}{\method~\textbf{(Ours)}} & \textbf{22.03} & \textbf{39.03}  & \textbf{55.06} \\
    \bottomrule
\end{tabular}

    }
    \caption{\textbf{Evaluation on \dataset cloudy-cloudy dataset.} Relative pose estimation results for visible-thermal image pairs taken under cloudy weather conditions.}
    \label{tab:cloudy-cloudy}
    \vspace{-.5cm}    
\end{table}

\subsection{Implementation Details}

We pre-train our model on the KAIST Multispectral Pedestrian Detection \cite{hwang2015multispectral} dataset, containing 95,000 visible-thermal pairs from a moving vehicle, with a focus on the top 640$\times$480 region to avoid road-dominant lower parts. We use Adam for pre-training with an initial learning rate of $2.5\times10^{-4}$ and a batch size of 2 for 9 epochs, taking 24 hours on 2 A5000 GPUs. For fine-tuning, we use the MegaDepth \cite{li2018megadepth} dataset with a 16 batch size at $640\times640$ resolution for padded images, employing Adam with a $2\times10^{-3}$ learning rate, converging after 24 hours on 8 A100 GPUs. Augmentation is applied randomly to one of the images $I^A$ or $I^B$. $\theta_c$, $\theta_f$ thresholds: $0.3$, $0.1$ respectively.

\begin{table}[htb]
    \centering
    \scriptsize{
        \setlength\tabcolsep{4.0pt}
\begin{tabular}{clcccc}
    \toprule
    \multirow{2}{*}[-.4em]{Category} & \multicolumn{1}{c}{\multirow{2}{*}[-.4em]{Method}} & \multicolumn{3}{c}{Pose estimation AUC}  \\
    \cmidrule(lr){3-5}
                                               &                                    & @5\degree & @10\degree & @20\degree           \\
    \midrule
    \multirow{3}{1.5cm}[-.0em]{Detector-based} &
     D2-Net~\cite{dusmanu2019d2}+NN             & \01.13    & \03.66      & \08.96     \\
                                               & SP~\cite{detone2018superpoint}+SuperGlue~\cite{sarlin2020superglue}               & \04.06     & 10.70      & 19.91     \\
                                               & SP~\cite{detone2018superpoint}+LightGlue~\cite{lindenberger2023lightglue} &  \03.12     & \08.39      & 15.29 \\
                                               & ReDFeat~\cite{deng2022redfeat} &  \01.16     & \03.24      & \07.22 \\
    \midrule
    \multirow{5}{1.5cm}{Detector-free}
                                               & \multicolumn{1}{l}{LoFTR~\cite{sun2021loftr}} & \02.77     & \07.71      & 16.36    \\
                                               & \multicolumn{1}{l}{LoFTR-MTV~\cite{liu2022multi}} & \00.92     & \02.79      & \06.82     \\
                                               & \multicolumn{1}{l}{ASpanFormer~\cite{chen2022aspanformer}} & \03.18 & \07.13  & 14.01 \\
                                               & \multicolumn{1}{l}{DKM~\cite{edstedt2023dkm}} & \0\underline{7.26} & \underline{14.63}  & \underline{23.60} \\
                                               & \multicolumn{1}{l}{\method~\textbf{(Ours)}} & \textbf{12.59} & \textbf{27.90}  & \textbf{45.03} \\
    \bottomrule
\end{tabular}

    }
    \caption{\textbf{Evaluation on \dataset cloudy-sunny dataset.} Relative pose estimation results for visible-thermal images taken under the cloudy and the sunny weather conditions.}
    \label{tab:cloudy-sunny}
    \vspace{-.5cm}
\end{table}

\subsection{Experiment 1: Relative Pose Estimation}
\PAR{Evaluation protocol:} To evaluate our method with the \dataset , following \cite{sun2021loftr}, we assess pose error using area under curve (AUC) at 5°, 10°, and 20° thresholds, defined as the maximum angular deviation from GT in rotation and translation. We employ RANSAC and a 1.5 threshold to solve for the essential matrix with predicted matches, setting the longer image side to 640 pixels during testing. Evaluations were independently conducted for cloudy-cloudy and cloudy-sunny sets.

\PAR{Compared methods:} We compared our \method  with the following publicly available methods: (1) Detector-based methods including D2-Net~\cite{dusmanu2019d2}, SuperGlue~\cite{sarlin2020superglue}, LightGlue~\cite{lindenberger2023lightglue} and ReDFeat~\cite{deng2022redfeat} , and (2) detector-free matchers including LoFTR~\cite{sun2021loftr}, LoFTR-MTV~\cite{liu2022multi}, ASpanFormer~\cite{chen2022aspanformer} and DKM~\cite{edstedt2023dkm}. LoFTR-MTV~\cite{liu2022multi} is LoFTR trained with aerial visible-TIR image pairs. Prior work in multi-modal image matching often lack public code or aren't readily benchmarked as multi-modal baselines.

\begin{table}[htb]
    \centering
    \scriptsize{
        \setlength\tabcolsep{4.0pt}
\begin{tabular}{clcccc}
    \toprule
    \multirow{2}{*}[-.4em]{Category} & \multicolumn{1}{c}{\multirow{2}{*}[-.4em]{Method}} & \multicolumn{3}{c}{Homography est. AUC}  \\
    \cmidrule(lr){3-5}
                                               &                                    & @5\degree & @10\degree & @20\degree           \\
    \midrule
    \multirow{3}{1.5cm}[-.0em]{Detector-based} &
     D2-Net~\cite{dusmanu2019d2}+NN             & \02.17    & \06.10      & 16.85    \\
                                               & SP~\cite{detone2018superpoint}+SuperGlue~\cite{sarlin2020superglue}               & \04.76     & \015.99      & 37.95     \\
                                               & SP~\cite{detone2018superpoint}+LightGlue~\cite{lindenberger2023lightglue} &  \05.57     & 15.83      & 35.42 \\
                                               & ReDFeat~\cite{deng2022redfeat} &  \03.72     & 12.13      & 29.21 \\
    \midrule
    \multirow{5}{1.5cm}{Detector-free}
                                               & \multicolumn{1}{l}{LoFTR~\cite{sun2021loftr}} & \06.34     & 14.22      & 30.23    \\
                                               & \multicolumn{1}{l}{LoFTR-MTV~\cite{liu2022multi}} & \04.56     & \08.57      & 16.23    \\
                                               & \multicolumn{1}{l}{ASpanFormer~\cite{chen2022aspanformer}} & \textbf{\09.50} & \underline{18.87}  & \underline{36.42} \\
                                               & \multicolumn{1}{l}{DKM~\cite{edstedt2023dkm}} & \02.79 & 9.55  & 25.87 \\
                                               & \multicolumn{1}{l}{\method~\textbf{(Ours)}} & \0\underline{8.19} & \textbf{23.37}  & \textbf{48.15} \\
    \bottomrule
\end{tabular}

    }
    \caption{\textbf{Homography estimation on LGHD LWIR/RGB \cite{lghd2015} and FusionDN \cite{xu2020fusiondn} datasets.} The AUC of the corner error is reported in percentage.}
    \label{tab:homography}
    \vspace{-.5cm}    
\end{table}

\PAR{Results:} As shown in Tab. \ref{tab:cloudy-cloudy} and \ref{tab:cloudy-sunny}, \method outperforms the other methods by a large margin in terms of relative pose estimation, demonstrating \method's effectiveness. When examining the LoFTR-MTV model trained on real \textbf{aerial} thermal and visible band images, it becomes evident that the proposed training approach is crucial for achieving accurate matching in urban settings. When we compare Tab. \ref{tab:cloudy-cloudy} and \ref{tab:cloudy-sunny}, we observe a decrease in the performance of our method due to the varying weather conditions. The qualitative results in Fig. \ref{fig:qulitative} support the quantitative results.

\subsection{Experiment 2: Homography Estimation}

\begin{figure}[bp]
    \vspace{-.5cm}
    \centering
    \includegraphics[width=.9\linewidth]{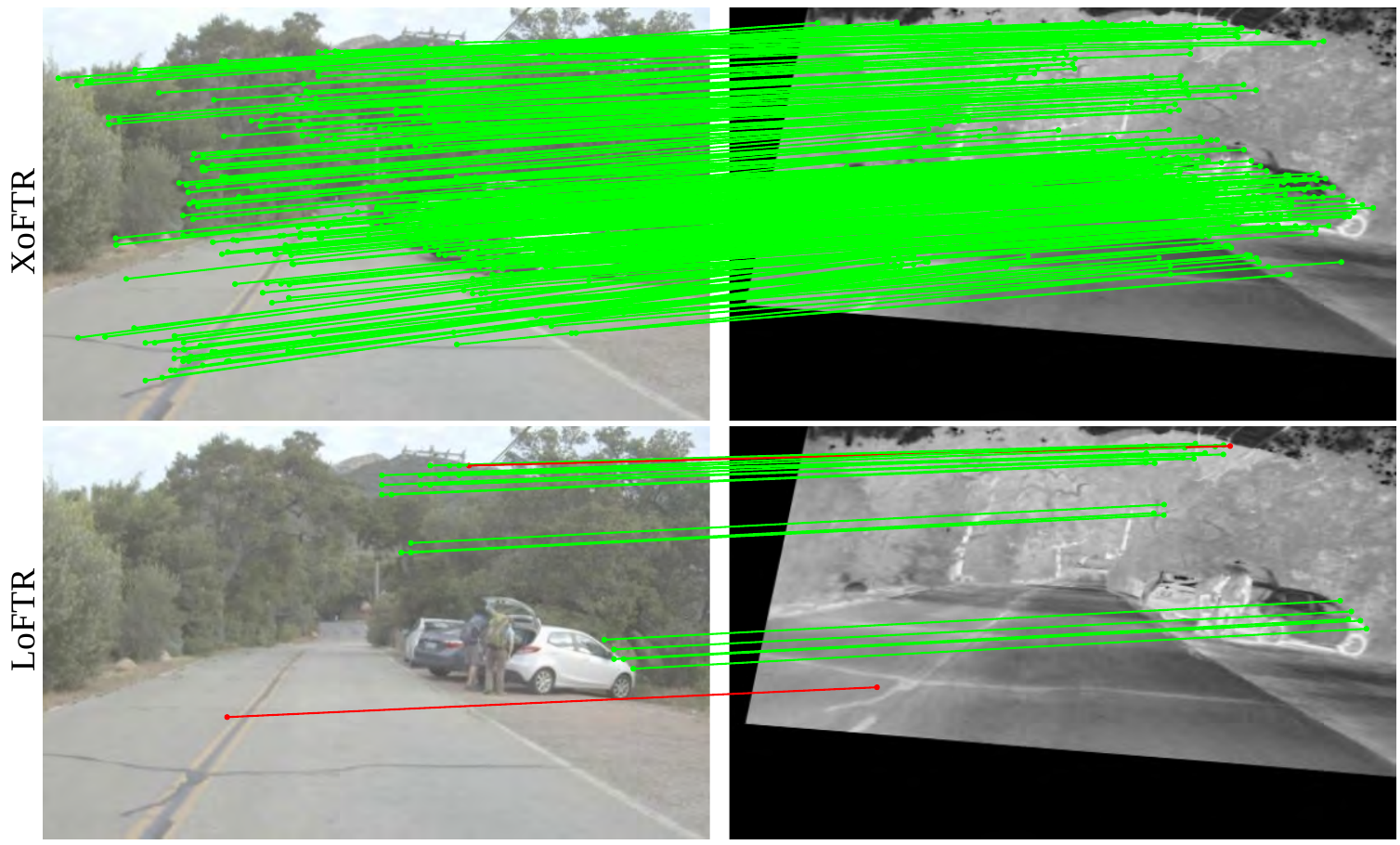}
    \caption{The qualitative homography estimation results for \method and LoFTR \cite{sun2021loftr}. }
    \label{fig:hom}
\end{figure}

\PAR{Dataset:} We utilized the LGHD LWIR/RGB \cite{lghd2015} dataset along with the RoadScene dataset from FusionDN \cite{xu2020fusiondn}, which contains 221 pairs of aligned visible-infrared images of road scenes with vehicles and pedestrians. The LGHD LWIR/RGB dataset comprises 44 pairs of aligned visible-thermal images of buildings. By merging these datasets, we obtained a new dataset. For each image pair, we generated 5 different homographies to serve as ground truth (GT) and applied them to the images, yielding a total of 1325 image pairs. Generated GT homographies include random scaling of $[0.8, 1.2]$, random perspective distortion $[-0.15, +0.15]$, and random rotation $[-15, +15]$ degrees.
\PAR{Evaluation protocol:} 
We used the same evaluation protocol that LoFTR uses for the HPatches \cite{hpatches_2017_cvpr} dataset, presenting results in terms of the area under the cumulative curve (AUC) for corner error distances of 3, 5, and 10 pixels.

\PAR{Results:}  Tab. \ref{tab:homography} demonstrates that \method surpasses other baseline methods across for 10 and 20 pixel error thresholds by a notable margin while Aspanformer \cite{chen2022aspanformer} gets the best result for the 5-pixel error threshold. Notably, the performance disparity between LoFTR and alternative approaches widens as the correctness threshold increases. This experiment confirms that our model performs well beyond our dataset, demonstrating its success in various situations. For qualitative comparison, see Fig. \ref{fig:hom}.

\subsection{Experiment 3: Ablation Study}
\label{sec:ablation}

We assess five different variants of \method evaluated on our \dataset cloudy-sunny dataset (Tab. \ref{tab:ablation}). The results suggest that: (1) Training from scratch for only image matching w/o pretext task yields an AUC drop as expected. (2) Fine-tuning w/o augmentation leads to a significant drop in AUC, showing the effectiveness of our proposed augmentation method. (3) Allowing only one-to-one assignment in coarse matching as in LoFTR results in a considerable drop in AUC, demonstrating the importance of one-to-many assignment. (4) Replacing our FLLM and CLMM with LoFTR's coarse-to-fine module (with one-to-one assignment) leads to a serious drop in AUC, showing the effects of the methods we use when bringing coarse matches to sub-pixel resolution. (5) Matching at the $1/2$ scale w/o SPRM yields an AUC drop due to imprecise matches. (6) Removing the second thresholding ($\theta_f$), which filters low confidence matches at $1/2$ scale, lowers AUC. (7) Removing the absolute positional bias from FLMM results in a drop in AUC.

\PAR{Running Time} On an A5000 GPU, \method runs 116 ms at $640\times512$ resolution while LoFTR \cite{sun2021loftr} runs 102 ms: One-to-many assignment  increases the number of the coarse matches leading to a small amount of process time increase.

\begin{table}[tb]
    \centering
    \scriptsize{
        \setlength\tabcolsep{4.0pt}
\begin{tabular}{lcccc}
    \toprule
    \multicolumn{1}{c}{\multirow{2}{*}[-.4em]{Method}} & \multicolumn{3}{c}{Pose estimation AUC}  \\
    \cmidrule(lr){2-4}
                                                & @5\degree & @10\degree & @20\degree           \\
    \midrule

                                               (1) without pretraining  & 11.81     & 26.51      & 42.93     \\
                                               (2) without augmentation &  \02.92     & \07.43      & 14.94 \\
                                               (3) with only one-to-one assignment & \09.95     & 22.60      & 38.73    \\
                                               (4) with coarse-to-fine module of LoFTR & \06.31     & 14.46      & 26.77     \\
                                               (5) without SPRM &  12.31     & 27.39      & 44.68 \\
                                               (6) without the second thresholding $\theta_f$ &  11.58     & 26.08      & 43.56 \\
                                               (7) without the positional bias in FLMM 
                                               &  12.23     & 26.99      & 43.36 \\
                                               \textbf{Full} \textbf{(\method)}             & \textbf{12.59}    & \textbf{27.90}      & \textbf{45.03}     \\
                                               
    \bottomrule
\end{tabular}
    }
    \caption{\textbf{Ablation study of \method.} All variants of \method are evaluated on \dataset cloud-sunny for pose estimation.}
    \label{tab:ablation}
    \vspace{-.5cm}
\end{table}

\section{Conclusion}
\label{sec:conc}


We have introduced \method as a novel pipeline for cross-view visible-thermal image matching. Our two-stage approach significantly outperforms the compared methods on several benchmarks. To better evaluate methods, we have also introduced a novel challenging dataset.


\PAR{Acknowledgements:} This work is funded by ROKETSAN Inc. The numerical calculations were partially performed at TÜBITAK ULAKBIM, TRUBA.

{
    \small
    \bibliographystyle{ieeenat_fullname}
    \bibliography{main}
}


\end{document}